\documentclass[10pt]{article} % For LaTeX2e
\pdfoutput=1
\usepackage[preprint]{tmlr}

\usepackage{amsmath,amsfonts,bm}

\def\eqref#1{equation~\ref{#1}}
\def\1{\bm{1}}

\DeclareMathAlphabet{\mathsfit}{\encodingdefault}{\sfdefault}{m}{sl}
\SetMathAlphabet{\mathsfit}{bold}{\encodingdefault}{\sfdefault}{bx}{n}

\usepackage{hyperref}
\usepackage{url}
\usepackage{graphicx}
\usepackage{grffile}
\usepackage{amsthm}

\newtheorem*{theorem*}{Theorem}

\title{Intriguing Properties of Modern GANs}

\author{\name Roy Friedman \email roy.friedman@mail.huji.ac.il \\
      \addr Department of Computer Science\\
      Hebrew University of Jerusalem
      \AND
      \name Yair Weiss \email  \\
      \addr Department of Computer Science\\
      Hebrew University of Jerusalem
      }

\newcommand{\lrpar}[1]{\left(#1\right)}
\newcommand{\lrbra}[1]{\left[#1\right]}

\def\month{MM}  % Insert correct month for camera-ready version
\def\year{YYYY} % Insert correct year for camera-ready version
\def\openreview{\url{https://{\usepackage{todonotes}}
openreview.net/forum?id=XXXX}} % Insert correct link to OpenReview for camera-ready version

\begin{document}

\maketitle

\begin{abstract}
Modern GANs achieve remarkable performance in terms of generating realistic and diverse samples. This has led many to believe that ``GANs capture the training data manifold''. In this work we show that this interpretation is wrong. We empirically show that the manifold learned by modern GANs does not fit the training distribution: specifically the manifold does not pass through the training examples and passes closer to out-of-distribution images than to in-distribution images.  We also investigate the distribution over images implied by the prior over the latent codes and study whether modern GANs learn a density that approximates the training distribution. Surprisingly, we find that the learned density is very far from the data distribution and that GANs tend to assign higher density to out-of-distribution images. Finally, we demonstrate that the set of images used to train modern GANs are often not part of the typical set described by the GANs' distribution.

% This has led many to believe in either of the following interpretations: (1) GANs learn a low dimensional manifold that fits the training distribution and (2) they learn a density that approximates the training distribution.  In this work we show that both of these interpretations are wrong. We empirically show that the manifold learned by GANs does not fit the training distribution (and not even the training examples). We go on to show that the learned density is very far from the data distribution, outperformed by other generative models. Finally, we show that the distributions learned by GANs are biased towards low-contrast images and that the set of images used to train GANs are often not part of the typical set of the GAN.

\end{abstract}

\section{Introduction}

Starting with their original formulation,  Generative Adversarial Networks (GANs,~\citealt{goodfellow2014generative}) quickly gained enormous popularity. As their popularity increased, so did the approaches to evaluating the quality of the images created by GANs and their fidelity to the true distribution (e.g. \citealt{heusel2017gans,richardson2018gans,arora2018gans,kynkaanniemi2019improved}). These methods exposed many problems in the distributions of images generated by GANs, such as mode collapse (\citealt{goodfellow2016nips} and artifacts within generated images (\citealt{karras2021alias}). 
%\todo{I added this sentence with the theorem that GANs are supposed to learn the true distribution}
Such developments were carried out with the understanding that ``the GAN's training criteria has a global minimum at the true distribution'' (from Theorem 1 of~\citeauthor{goodfellow2014generative}).

Following these evaluations, there was a refinement of GANs and such failures have become much less prominent. 
For example, FID (\citealt{heusel2017gans}) and recall (\citealt{kynkaanniemi2019improved}) both check whether the full diversity of the training distribution is captured by the generative model. GANs which have both low FID and high recall are deemed not to suffer from mode collapse, as they supposedly capture the full support of the training data.
Due to their impressive performance in such metrics, many applications have progressed with a working assumption that GANs learn the ``training data manifold'' (\citealt{schlegl2017unsupervised})
%\todo{Now it is an actual quote}.
This assumption has even carried through to domains where an incorrect model of the true distribution can incur heavy risks, such as super-resolving MRI scans (\citealt{bendel2022regularized}), simulations for self-driving cars (\citealt{zhang2018deeproad}), detecting cancer through GAN inversions (\citealt{schlegl2017unsupervised}), recovering frequencies in astrophysical images (~\citealt{schawinski2017generative}) and more.

In this work we study GANs with excellent performance and question whether these GANs have actually learned the true data manifold. By critically analyzing GANs through this lens, we find that they behave in ways counter to popular belief.

As mentioned above, modern GANs are frequently thought of as manifold methods, wherein the GANs' generator captures the image manifold. Surprisingly, we find that training examples are not part of the learned manifold. We go on to demonstrate that the GAN manifold passes almost the same distance from the class of images the generator was trained on as other classes, even when measured with a perceptual distance (\citealt{zhang2018unreasonable}). This failure is intriguing as a GAN trained on ships, for instance, is only able to generate images of ships and not other object classes. 
Furthermore, we find that the GAN manifold passes unusually close to images outside the training domain. In particular, we show that the GAN manifold passes closer to SVHN images than to the images it was trained on. 

While GANs are most commonly viewed as manifold methods, we demonstrate how analyzing them only in these terms might not capture the full breadth of their capabilities. In particular, the manifold view ignores the density the generator assigns to different parts of space and whether or not the manifold is more abundant in regions around the true distribution. To account for this, we analyze the density the GAN's generator assigns to samples in terms of the average test log-likelihood. While mostly overlooked in the GAN literature, the log-likelihood is widely used to evaluate other families of generative models, such as score based models (\citealt{song2020score}) and normalizing flows (\citealt{kingma2018glow}). 
Moreover, the log-likelihood was used in the past to show that large families of generative models assign higher likelihood to images outside the training distribution than those used in training (\citealt{nalisnick2018deep}) and that many conditional generative models are weak classifiers (\citealt{fetaya2019understanding}). Works such as those by~\citeauthor{nalisnick2018deep} focused on models where the likelihood is tractable, which have much worse FID scores than modern GANs. There is scarce literature of this vein concerning generative models that show excellent results on metrics other than average test log-likelihood.

By evaluating GANs with low FID as density estimators, we show that modern GANs under-perform in terms of average test log-likelihood when compared to models with much worse FID. We go on to show that GANs are biased towards images with larger ``flat'' areas. Indeed, the log-likelihood assigned by the GAN is anti-correlated with local variance in the image.
Finally, we question whether the images used to train the GAN are part of the typical set of images, a test previously used to determine whether a set of images are outliers according to a generative model (\citealt{nalisnick2019detecting}). %e find that the training examples are \emph{not} part of the typical set of the GAN; that is, the probability that the GAN would generate this set of images is essentially zero.

To summarize, in this paper we discuss four intriguing properties of modern GANs:
\begin{itemize}
    \item The learned manifold does not pass through the training examples.
    \item The learned manifold is closer to out-of-distribution images than to in-distribution images.
    \item The density model learned by by the GAN assigns higher density to out-of-distribution images. 
    \item The training images are not in the typical set of the GAN. 
\end{itemize}

\begin{figure}
\begin{center}
    \includegraphics[width=.8\linewidth]{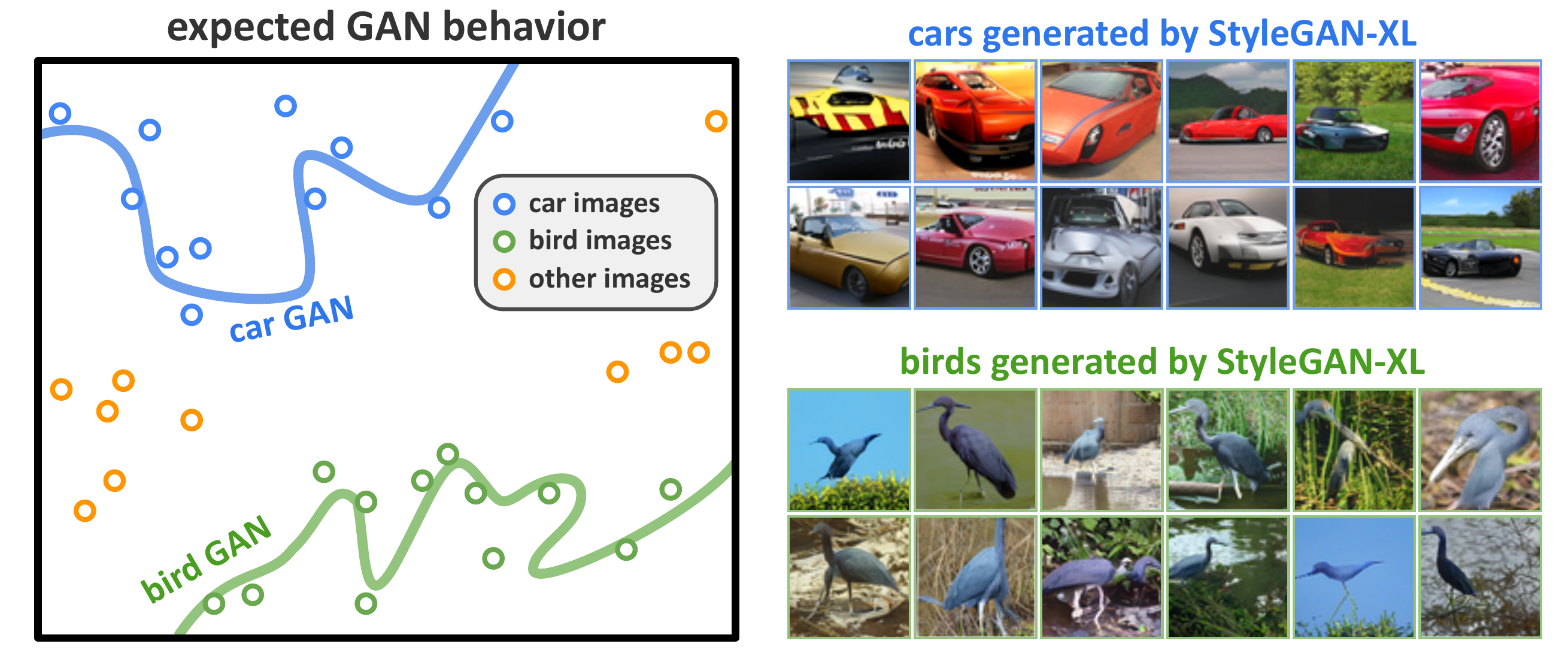}
\end{center}
  \caption{Modern GANs work amazingly well, to the point that it is expected that they capture the data manifold (left). It is typically assumed that because GANs can generate realistic images (right) that they capture the true data manifold. In this paper we show that this assumption is false.}
  \label{fig:expected-behavior}

\end{figure}

\section{Background}

\begin{figure}
    \centering
    \includegraphics[width=\textwidth]{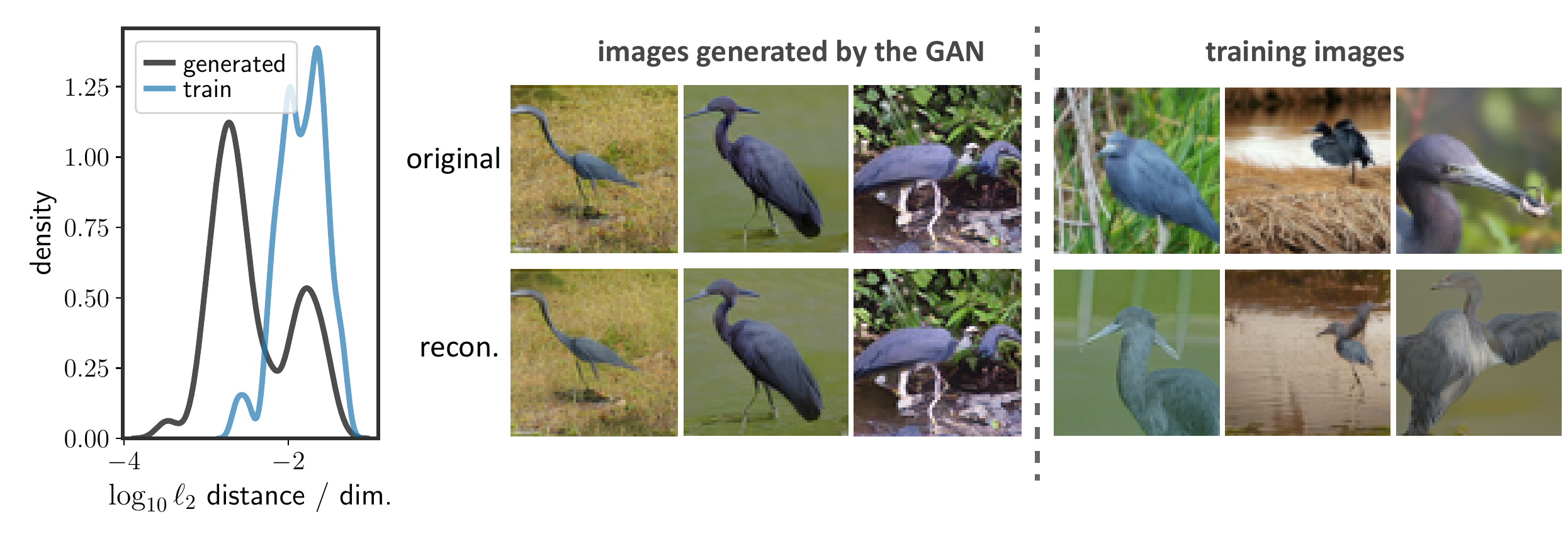}
    \caption{
    \textbf{Left:} $\ell_2$ distance (lower is better) of reconstructions of generated and train images under StyleGAN-XL (\protect{\citealt{sauer2022stylegan}}) trained on ImageNet (when only $z$ is optimized). The distribution of reconstruction errors for training images is distinct from the same for images generated by the GAN, implying that the training images are not part of the GAN manifold. \textbf{Right:} examples of projections to the GAN. Generated images are part of the manifold (and can be reconstructed), but training images are not. For training images, the reconstructed images are realistic birds but they are different birds.}
    \label{fig:train-recon}

\end{figure}

\subsection{Evaluation of Generative Models}
State of the art GANs (\citealt{sauer2022stylegan,karras2020analyzing,kang2023scaling}) are typically evaluated by methods that analyze the statistics of the generated samples (\citealt{heusel2017gans,kynkaanniemi2019improved,richardson2018gans}). Almost all of these approaches define a statistic that is zero only if the GAN-generated and true samples are from the same distribution. Methods such as these are extremely effective in determining whether the generative model has learned the correct statistic but are somewhat limited. One such limitation is that for data in very high dimensions, many samples might be needed to determine if the true and generated distribution are distinct. Another is that the difference between the distributions is only measured on particular statistics, ignoring the others.

Many generative models other than GANs utilize the average test log-likelihood (LL) as an additional evaluation metric, since high test LL is equivalent to low KL divergence between the generative distribution and the true data distribution. For example, score-based models (\citealt{song2020score}) and normalizing flows (\citealt{kingma2018glow,nalisnick2018deep,fetaya2019understanding}) report the test LL or the negative test LL as a way to compare the quality of the model. Since the LL is never explicitly used in GANs, it was largely ignored when comparing such models and calculated in only relatively few studies for small scale models on simple images, such as MNIST (\citealt{wu2016quantitative,theis2015note}).

In the past, generative models were also evaluated through their performance in inference tasks, such as generative classification (e.g. \citealt{fetaya2019understanding,hinton1995wake}). In generative classification, a generative model $p_{\theta_c}(x|c)$ is separately trained for each class $c$ in the data. An unseen sample is then classified by determining which class-conditional LL is the largest:
\begin{equation}
    \hat{c}(x)=\arg\max_c \log p_{\theta_c}(x|c)
\end{equation}
where $\hat{c}(x)$ is the predicted class for sample $x$. If $p_{\theta_c}(x|c)\approx p_{\text{data}}(x|c)$ for all classes $c$, then the above classification scheme is guaranteed to be optimal (\citealt{duda1973pattern}). On the flip side, if the classification performance is sub-optimal, then in no uncertain terms $p_{\theta_c}(x|c)\neq p_{\text{data}}(x|c)$.
Using inference tasks to evaluate generative models can expose biases in the generative distribution, and is potentially a sample-efficient way to find out whether a generative model hasn't learned the true data distribution.

\subsection{Annealed Importance Sampling}

Calculating the LL of any decoder-based model involves calculating the following integral:
\begin{equation}\label{eq:integral}
    p_\theta(x)=\intop p_\theta(x|z)p(z)dz
\end{equation}
where $p(z)$ is the prior over the latent dimension and $p_\theta(x|z)$ is the probability of generating $x$, using the model, with the latent code $z$. Unfortunately, analytically solving this integral is intractable for complex decoder-based models. Instead, approximate methods such as Markov Chain Monte Carlo (MCMC) must be used in order to calculate the LL. In particular, \citeauthor{wu2016quantitative} showed that annealed importance sampling (AIS,~\citealt{neal2001annealed}) can be used to accurately approximate the LL of GANs. 

AIS is an MCMC approach that uses multiple intermediate distributions in order to estimate the integral in equation~\ref{eq:integral}. \citeauthor{wu2016quantitative} showed that an accurate estimate can be achieved, using many intermediate distributions (i.e. long MCMC chains) and multiple chains. When used to calculate the \emph{log} of equation~\ref{eq:integral}, the estimate is only a stochastic lower-bound (\citealt{grosse2015sandwiching}) of the LL that gets tighter as more intermediate distributions are introduced. For a more detailed explanation, see appendix~\ref{a:ais} or~\citeauthor{wu2016quantitative}.

\section{Are GANs Good Manifold Methods?}

GANs transform samples from a low dimensional latent space, $\mathcal{Z}$, into the larger dimension of the data $\mathcal{X}$, effectively describing a manifold. In much of the literature, it is believed that the GAN captures the true low dimensional behavior of the training data (e.g.~\citeauthor{menon2020pulse,schlegl2017unsupervised,xiao2021tackling,bendel2022regularized,schawinski2017generative}). A simple schematic of this assumption can be seen in figure~\ref{fig:expected-behavior}, where it is usually expected that the images from the training distribution are very close to, or exactly on, the learned manifold. Furthermore, points that are not part of the training distribution are expected to be far from the learned manifold, whether they are natural images or not.

If GANs truly behave as believed and shown in figure~\ref{fig:expected-behavior}, it should be possible to use them for classification and outlier detection (\cite{schlegl2017unsupervised}).
% , as shown in figure~\ref{fig:manifold-inference}. 
For classification, an image is separately projected onto GANs trained on each of the classes and is classified according to the manifold whose projection was closest to the original image. Outlier detection can be carried out analogously, where the point is projected to the GAN manifold and if it's distance from the original point is too large it is deemed an outlier.

\subsection{Inference with the GAN Manifold}
\begin{figure}
\begin{center}
    \includegraphics[width=.8\linewidth]{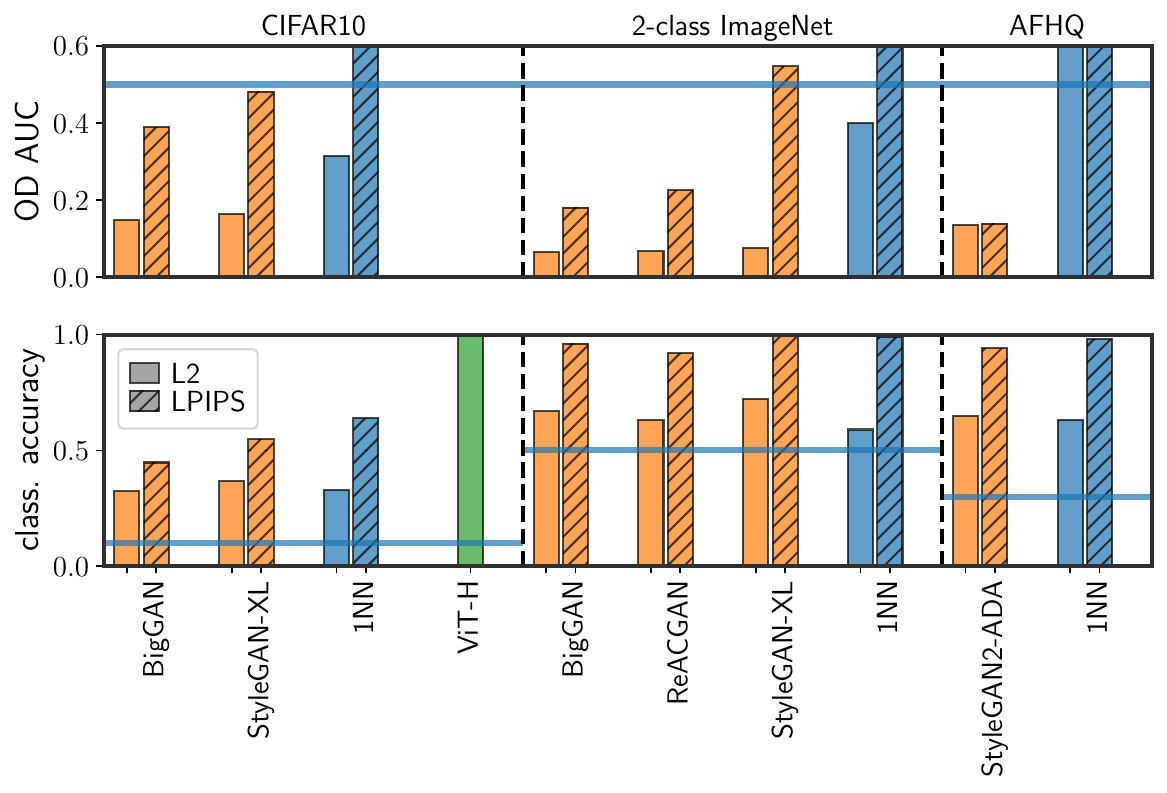}
\end{center}
  \caption{Performance of different GANs at the task of outlier detection (top) and classification (bottom). In all cases, both the $\ell_2$ (plain) and the LPIPS (hatched, \protect{\citealt{zhang2018unreasonable}}) distances are used. The GANs are compared to a 1-nearest neighbor (1NN) baseline. For CIFAR10, all methods are also compared with ViT-H (\protect{\citealt{dosovitskiy2020image}}). The GANs always underperform, compared to the 1NN baseline.}
  \label{fig:manifold-performance}

\end{figure}
\begin{figure}
\begin{center}
    \includegraphics[width=\linewidth]{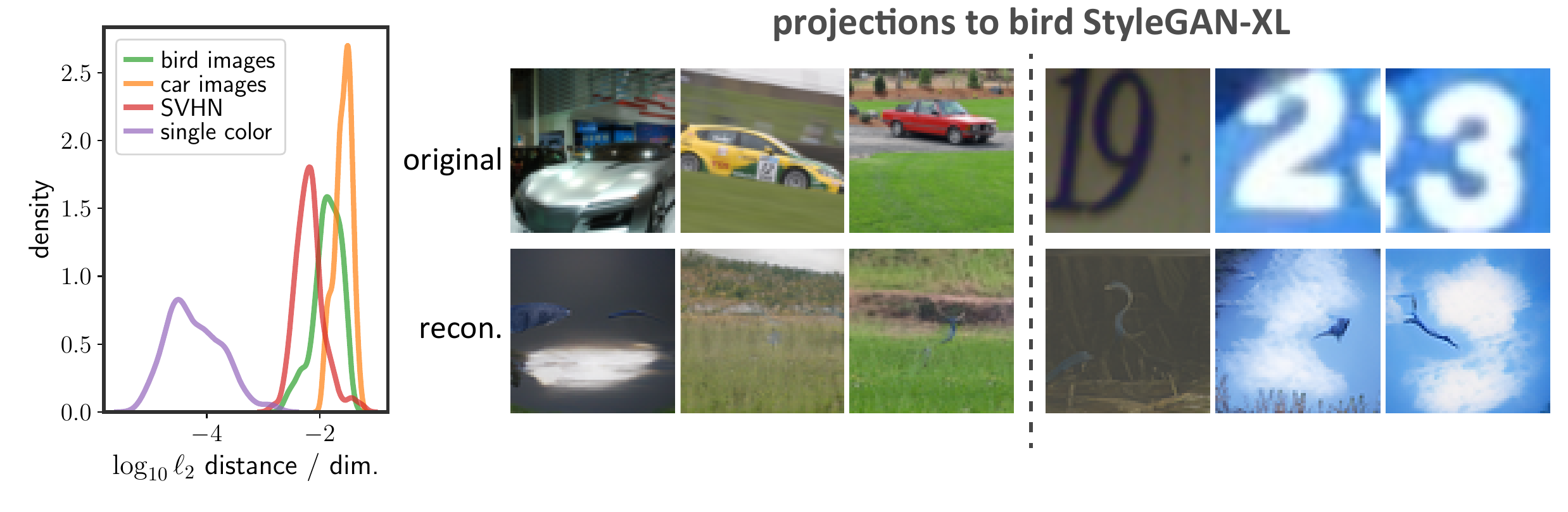}
\end{center}
  \caption{\textbf{Left:} $\ell_2$ reconstruction errors (lower is better) for the reconstruction of different image groups by StyleGAN-XL trained on birds. The reconstruction error for birds or cars is similar, while there exist images that the GAN can reconstruct much better than those it was trained on. \textbf{Right:} examples for StyleGAN-XL reconstructions of car images and rescaled SVHN images. In all cases, the GANs capture the overall image structure, but don't retain the identity of the main object in the image, leading to projections which are far from the image whether in the training domain or not.}
  \label{fig:inversions}

\end{figure}

We test whether modern GANs can be considered as manifold methods using a range of GANs and various datasets. The GANs we use are: StyleGAN-XL (\citealt{sauer2022stylegan}) with an FID of 1.94 on ImageNet and 1.88 on CIFAR10, ReACGAN (\citealt{kang2021rebooting,kang2022studiogan}) with an FID of 8.19 on ImageNet and 3.87 on CIFAR10, BigGAN-DiffAug (\citealt{zhao2020differentiable}) with an FID of 8.7 on CIFAR10, BigGAN (\citealt{kang2022studiogan,brock2018large}) with an FID of 8.54 on ImageNet, and StyleGAN2-ADA (\citealt{karras2020training}) with an FID of 3.55 on AFHQCat and 7.4 on AFHQDog. To project the images on the GAN, we optimize the latent $z$ code from multiple restarts and choose the optimized code with the smallest reconstruction error. Further details can be found in appendix~\ref{a:details}. All of the GANs chosen are able to generate highly realistic images and seemingly capture the full diversity of the datasets they were trained on.

In the task of classification, we analyze the performance of the GANs on a subset (~1000 test samples) of the CIFAR10 dataset, a binary problem from ImageNet (``sports car'' class versus the ``blue heron'' class) and a subset (~1000) of AFHQ images, where the classes are AFHQDog, AFHQCat and AFHQWild.
For outlier detection in all datasets, we consider images of a single color and rescaled versions of SVHN images as outliers and report the area under the ROC curve (OD AUC).

The first assumption we question is whether the GAN manifold actually passes through the training examples. For a given input $x$ we can determine if it is on the GAN manifold by using gradient descent to approximately solve the optimization problem:
\begin{equation}
z^* = \arg\min_z \|G(z)-x\|
\end{equation}

Surprisingly, it turns out that most of the training points are \emph{not part of the GAN manifold}. Figure~\ref{fig:train-recon} (right) shows that for samples generated by the GAN, the optimization algorithm succeeds in finding a latent $z$ so that the generated images is almost identical to the input image $x$. But for images from the training set, the best $z$ found by the optimization algorithm generates an image that looks very different, although realistic. 
The figure also shows histograms of the reconstruction error of training samples found by inverting them from the $\mathcal{Z}$ space for StyleGAN-XL on ImageNet (for results on more GANs, see appendix~\ref{a:results}). As a baseline, samples generated by the GAN are also inverted and the histogram of their reconstruction error is shown as well. In general, the reconstruction error of training images is an order of magnitude worse than the reconstruction error of a generated image.

We then test the classification accuracy and outlier detection capabilities of the GANs. If the GANs have learned something close to the true data manifold, then they should act as good classifiers and outlier detectors. As a baseline, we compare the GANs to a simple 1-nearest-neighbor (1NN) estimator. Under the 1NN estimator, a sample is classified according to the class of its nearest neighbor. In outlier detection, a sample is considered an outlier if it is further than a pre-defined threshold from its nearest neighbor in the training distribution. This baseline is analogous to a GAN that has completely memorized the training examples and can't generate anything outside the training examples.

We again find that the learned manifolds do not behave as expected. Using the GAN manifolds for classification results in poor accuracy, as shown in figure~\ref{fig:manifold-performance} (bottom, in orange). The GANs are usually worse than the simple 1NN classifier (in blue), even when using a more perceptual distance such as LPIPS (\citealt{zhang2018unreasonable}). 
The performance of GANs as outlier detectors is not much better, as shown in figure~\ref{fig:manifold-performance} (top) - the performance of most GANs is worse than random, even when using the perceptual distance. On the other hand, the 1NN outlier detector is always better than chance when using LPIPS. 

We can better understand this poor performance by explicitly looking at the projections to the GAN manifold. Figure~\ref{fig:inversions} (left) shows the reconstruction errors that different image groups have under a StyleGAN-XL trained on ImageNet blue heron (bird) images. Many images of sports cars have the same reconstruction error as images of birds, explaining why GANs are so bad at classifying between different images. Moreover, the GAN is able to reconstruct images of a single color and sometimes those from SVHN better than images from the distribution they were trained on.
Figure~\ref{fig:inversions} (right) shows projections of images of sports cars (top) and SVHN (bottom) onto the bird GAN. The projection of any image keeps the overall image structure, even for images from the wrong class. Because of this, the projection's distance from the original image is about the same whether the original was from the training distribution or not. 

Such behavior is surprising. Images of sports cars and SVHN digits are frequently very different than those of blue herons, but the GAN is still able to reconstruct much of the context of the images. In images of cars, the GAN reconstruct the background, removing the car. For SVHN, the dominant colors and positions of the digits are kept. This is the same reconstruction behavior for images from the true distribution, an example of which can be seen in figure~\ref{fig:train-recon} (right) for training images.

\section{Are GANs Good Density Estimators?}
\begin{figure}
\begin{center}
    \includegraphics[width=.5\linewidth]{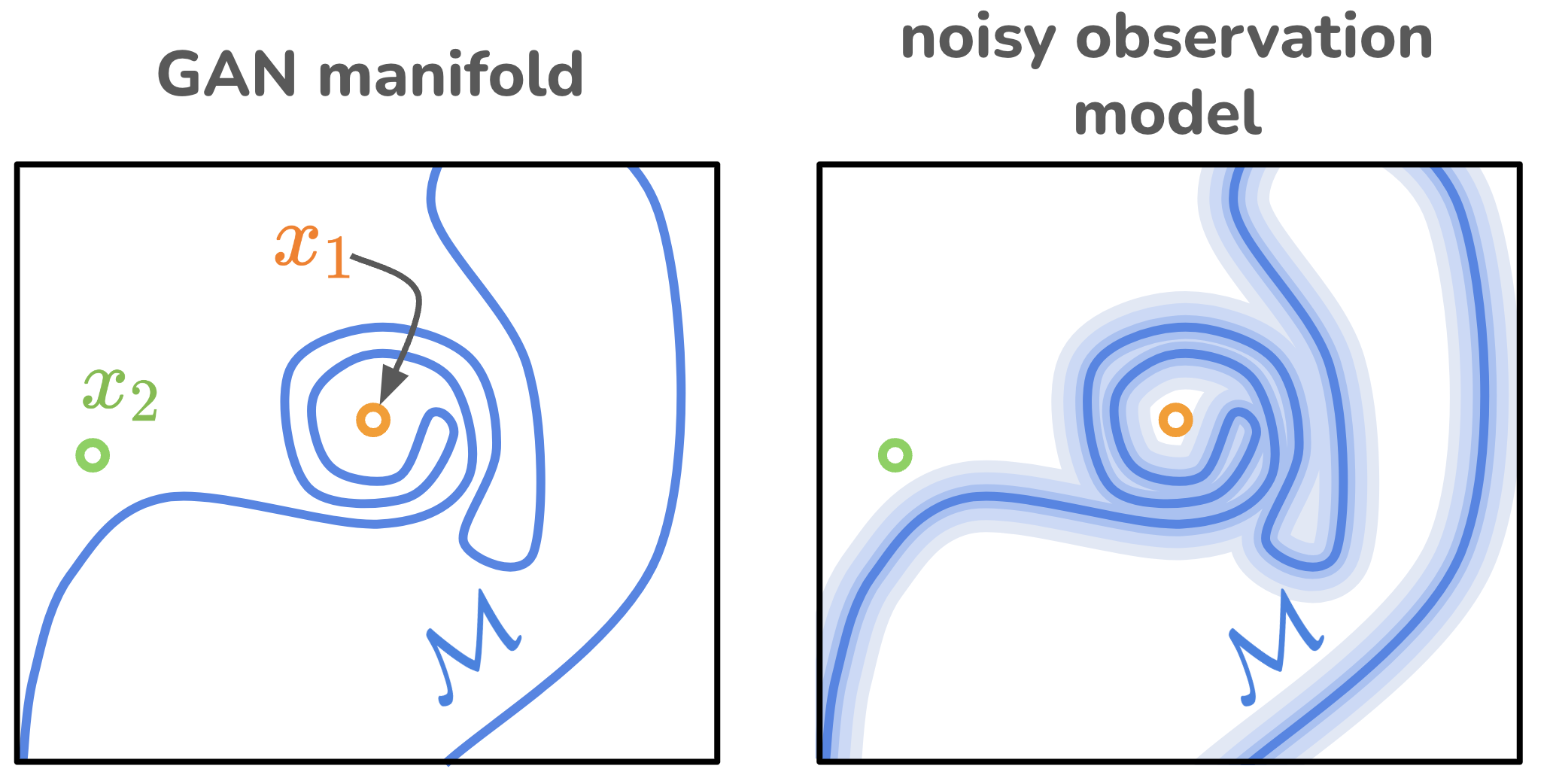}
\end{center}
  \caption{An example of a case where simply calculating the distance from the manifold might not tell the whole story. In this case, both $x_1$ and $x_2$ are the same distance from the manifold but the manifold is more abundant near $x_1$, making images in its region more plausible (left). By adding an observation model (right), queries to the model are transformed to calculating the log-likelihood under this observation model, essentially integrating over all possible areas of the manifold. With the observation model (right), $x_1$ is assigned a higher log-likelihood than $x_2$, even though they are equally distant from the manifold.}
  \label{fig:curved-manifold}

\end{figure}
\begin{table}
\centering
\resizebox{.6\linewidth}{!}{
\begin{tabular}{|r|l|c|c|}
\hline
Model & Dataset & LL (bits/dim.) $\uparrow$ & FID $\downarrow$ \\
\hline
\hline

BigGAN-DiffAug & CIFAR10 & -6.68 & 8.7 \\
StyleGAN-XL & CIFAR10 & -7.27 & 1.88 \\
Glow (\citealp{kingma2018glow}) & CIFAR10 & -3.35 & 3.35 \\
DDPM++ (\citealp{song2020score}) & CIFAR10 & -2.99 & 2.92 \\
\hline
\hline
BigGAN & ImageNet & -7.59 & 8.54 \\
StyleGAN-XL & ImageNet & -7.29 & 1.94 \\
Glow (\citealp{kingma2018glow}) & ImageNet & -3.81 & - \\
VDM (\citealp{kingma2021variational}) & ImageNet & -3.4 & - \\
\hline
\hline
StyleGAN2-ADA & AFHQ & -7.75 & 3.55 \\
\hline
\end{tabular}
}
\caption{GAN performance as a density estimator next to other generative models. Although GANs sometimes come ahead in terms of FID, their average test log-likelihood is always lower.}
\label{tab:LL-perf}
\end{table}

Viewing GANs purely as manifold methods potentially ignores two important aspects. First, the prior over the latent codes is completely ignored in all of the tests conducted above. Furthermore, while the GAN is equally  far from points in or outside of the distribution, a larger portion of it might pass closer to the training samples than to other samples. An example of this can be seen in figure~\ref{fig:curved-manifold} (left): while both points are equally  far from the manifold, more of the manifold passes close to the point in the middle of the spiral ($x_1$), and much less near the other point ($x_2$). Such behavior would explain why GANs are able to generate realistic images from the correct class while still being equally distant from points inside and out of the training distribution. To account for both of these aspects, we evaluate the GANs as density estimators, implicitly taking both the distribution over the latent space and the curvature of the manifold into account.

\subsection{Adding an Observation Model}
\begin{figure}
\begin{center}
    \includegraphics[width=.8\linewidth]{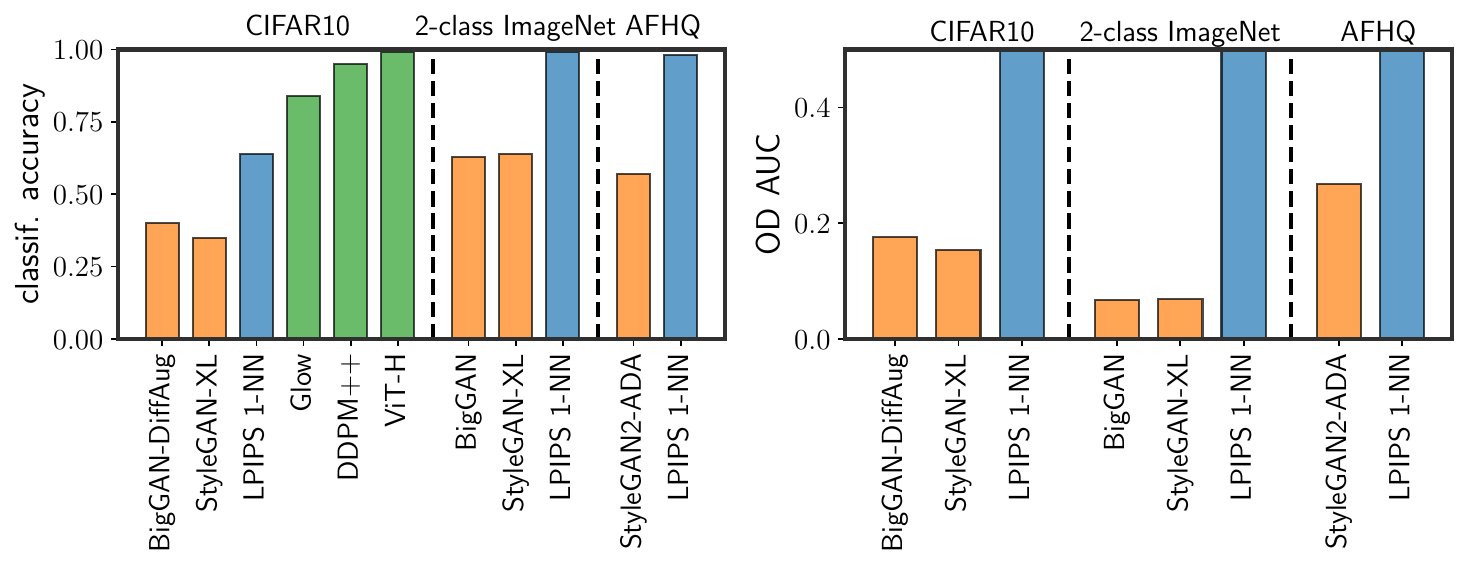}
\end{center}
  \caption{Performance of GANs as generative classifiers (left) and outlier detectors (right). The GANs are outperformed by simple baselines, as well as other generative models.}
  \label{fig:LL-perf}

\end{figure}

When distributions are defined on a low-rank space, it is common to assume an added ``observation noise'' (e.g. \citealt{wu2016quantitative,kingma2013auto,tipping1999principal}). In practice, this amounts to the generative model $x=G_\theta(z)+\text{noise}$ where the noise is often assumed to be Gaussian with a fixed variance, $\sigma^2$. The density (or likelihood) of a sample under the noisy model is then given by:
\begin{equation}
    p_\theta(x)=\intop p_\sigma\lrpar{x|\ G_\theta(z)}p(z)dz
\end{equation}
As mentioned by~\citeauthor{wu2016quantitative}, while the Gaussian observation model is possibly too simplistic, it can used to better understand the distributions learned by GANs. However, as long as this observation is used to compare GANs to each other, then we believe that it does give an indication of which GAN better captures the distribution. 

We use the above described observation model in order to estimate the density GANs assign to previously unseen samples, conducting the same experiments as those for the manifold.

In all of our experiments, we use AIS with 500 steps, 4 chains, a Hamiltonian Monte Carlo (HMC) kernel and 10 leap-frogs to estimate a sample's LL. These values were chosen to mitigate the high computational load of running AIS while still extracting accurate estimates of the LL. More experimental details can be found in appendix~\ref{a:ais-details}. The variance of the observation model was chosen to be the mean variance of the reconstruction error of training samples, which is the maximum likelihood estimator for the observation noise. In all cases, the PSNR of the images was high; that is, the variance of the noise assumed was much smaller than the strength of the signal. For more details regarding the variances used, see appendix~\ref{a:variance}.

\subsection{Evaluation as Density Models}

Using the added observation model described above, we calculate the LL of the different GANs on unseen data and compare them to other generative models. This comparison can be found in table~\ref{tab:LL-perf}. GANs assign much lower LL to test samples than other methods, even when their FID is better than the other models. 

Figure~\ref{fig:LL-perf} shows the performance of the GANs as generative classifiers (left) and outlier detectors (right), next to the LPIPS 1NN baseline and other generative models. The classification accuracy of Glow is as reported by~\citeauthor{fetaya2019understanding} and that of DDPM++ as reported by~\citeauthor{zimmermann2021score}.
Once again, the GANs are completely outperformed by the relatively simple 1NN estimator, indicating that the density GANs assign to samples is different than that of the true distribution.

\begin{figure}
\begin{center}
    \includegraphics[width=\linewidth]{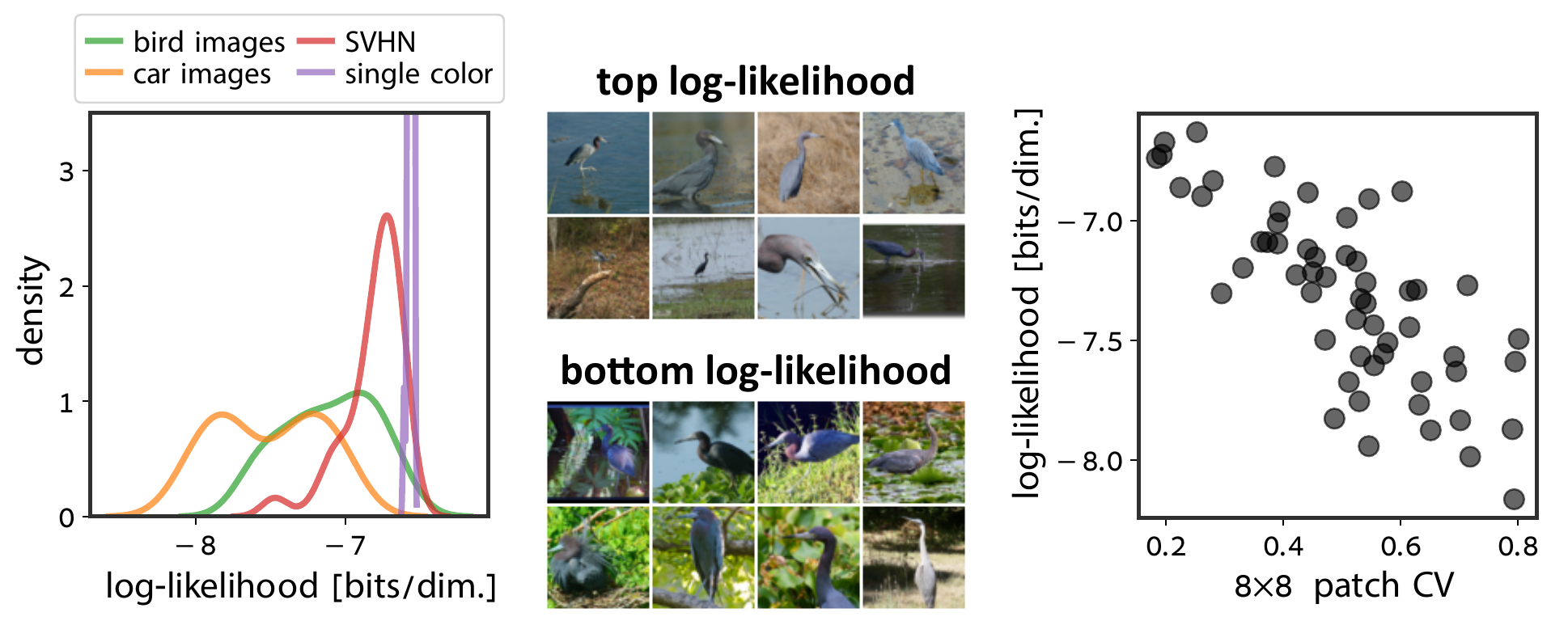}
\end{center}
  \caption{\textbf{Left:} log-likelihoods (higher is better) that different image groups attain under a StyleGAN-XL trained only on birds. Images of single color and SVHN images have higher log-likelihoods than images from the domain the GAN was trained on. \textbf{Middle:} the top and bottom images according to the log-likelihood. GANs tend to have a bias towards images with larger flat region. \textbf{Right:} a scatter plot of the images' log-likelihood against the average coefficient of variation (CV) of 8x8 patches. The log-likelihood assigned by the GAN is anti-correlated with the average variance of the patches in the image.}
  \label{fig:top-bottom-LL}

\end{figure}

Moreover, GANs tend to assign higher LL to images completely outside their training distribution, as can be seen in figure~\ref{fig:top-bottom-LL} (left): the StyleGAN-XL bird GAN gave higher LL to SVHN images than to test images. 

To further understand the behavior of the GANs as density estimators, we look at which images have the highest and lowest LLs. The images with highest and lowest LLs from the ``blue heron'' class of ImageNet can be seen in figure~\ref{fig:top-bottom-LL} (center). Visually it seems that GANs prefer images with larger flat areas and low contrast. To be specific, figure~\ref{fig:top-bottom-LL} shows how the average coefficient of variance in small 8x8 patches of the image is anti-correlated with the LL (on the ``sports car'' and ``blue heron'' classes of ImageNet).

% \subsection{Curvature of Modern GANs}

% \red{maybe I should remove this part?}

% As a bi-product of the AIS procedure, we can explicitly check whether or not the GAN's behavior is similar to the one shown in figure~\ref{fig:curved-manifold}; that is, whether a larger portion of the manifold passes close to images from the target distribution. 

% \red{Figure fig:KL-vs-recon - a figure that shows the breakup of the log-likelihood into the reconstruction and KL terms}

% The log-likelihood of a sample can be decomposed into a \emph{reconstruction term} and a \emph{KL term} (the derivation of this can be found in appendix~\ref{a:LL-KL-rec}):
% \begin{equation}\label{eq:ll-energy}
%     \log p_\theta(x)=\underbrace{\mathbb{E}_{z|x}\lrbra{\log p_\sigma \lrpar{x|\ G_\theta(z)}}}_{\text{recon. term}} - \underbrace{D_\text{KL}\lrpar{p_\theta(z|x)\ ||\ p(z)}}_{\text{KL term}}
% \end{equation}
% The reconstruction term is mostly affected by the actual reconstruction quality of the images, while the KL term is the one that changes if there are many possible latent codes that can generate close variations of the image. 

% Figure~\ref{fig:KL-vs-rec} shows the makeup of the log-likelihood of the different GANs on a wide set of images. In almost all cases, the reconstruction term is the one that dominates the changes in the log-likelihood of the images. That is, the curvature of the GAN is mostly constant throughout the whole space.

\subsection{Typicality of Training Examples}
While the samples from the true distribution might have lower LL than other sets of images, this alone is not an indication of a failure of the GAN. A stronger test is that of \emph{typicality} of a set of images. For the distribution $p_\theta(x)$, we follow \citeauthor{nalisnick2018deep} and define the $\lrpar{\epsilon, N}$-typical set $\mathcal{A}_\epsilon^N(p_\theta)$ of the distribution as all sets of $N$ samples that satisfy:
\begin{equation}
    \left|\frac{1}{N}\sum_{i=1}^N \log p_\theta(x_i) + H\lrpar{p_\theta}\right|\le \epsilon
\end{equation}
where $H\lrpar{p_\theta}$ is the entropy of the distribution $p_\theta(x)$. When $N$ is large, then (\citealt{cover1999elements}):
\begin{equation}
    P\lrpar{\mathcal{A}_\epsilon^N(p_\theta)}>1-\epsilon
\end{equation}
That is, when $N$ is large enough, this set of samples captures most of the distribution according to $p_\theta(x)$.

We begin by drawing i.i.d. samples from this typical set, i.e. we generate images from the GAN. We then calculate their LL using AIS, in exactly the same procedure as that for the training and SVHN images. The average LL of images generated by the GAN gives us an estimate of the entropy of $p_\theta(x)$: $\hat{H}\lrpar{p_\theta}=-\frac{1}{N}\sum_{i=1}^N\log p_\theta(x)$ where $x_i\sim p_\theta(x)$. $\epsilon$ is chosen such that in 95\% of the time, 50 samples drawn from the model are part of the typical set (i.e. the bootstrap confidence interval suggested by~\citeauthor{nalisnick2019detecting}).

\begin{figure}
\begin{center}
    \includegraphics[width=.9\linewidth]{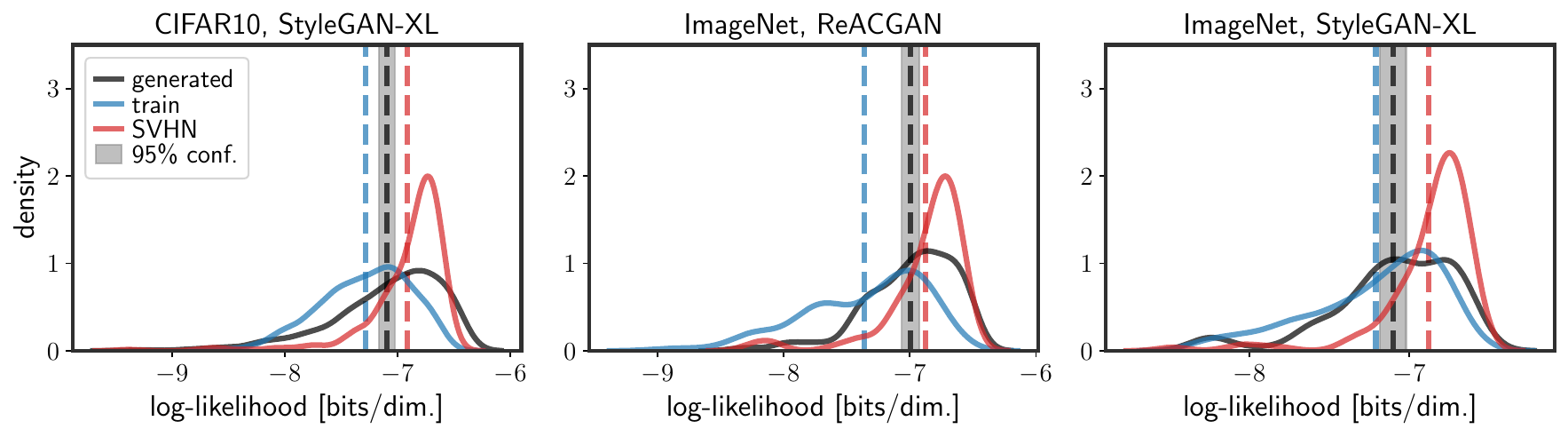}
\end{center}
  \caption{Histograms of log-likelihoods assigned to images generated by the GAN (black), train images (blue) and SVHN images (red). Each plot is for a different GAN-dataset pair. The dashed vertical lines are the means of the log-likelihood of the corresponding colored histogram, and the shaded area around the mean of the generated images are values of mean log-likelihoods associated with the typical set of the GAN. In all cases, the training images are outside this shaded area, meaning the set of training images \emph{is not} part of the typical set of any of the GANs.}
  \label{fig:typical-set}

\end{figure}

Using the typicality test suggested by~\citeauthor{nalisnick2019detecting}, we find that for the GANs we studied the set of \emph{training samples is not part of the typical set} of the distribution defined by the GAN. Sometimes images from SVHN are even closer to being part of the typical set of the GAN than training images. This can be seen in figure~\ref{fig:typical-set} where histograms of the LLs for images generated by the GAN are shown next to histograms for other images. The dashed vertical lines are the means of the histograms for each image type and the shaded region is the $\epsilon$-ball around the estimate of the entropy containing the typical sets $\mathcal{A}_\epsilon^N\lrpar{p_\theta}$, according to a bootstrap estimator of $\epsilon$. The mean LL for training images is always outside the shaded area, indicating that training images are not part of the typical sets of the GANs.

\section{Related Works}

This work follows a line of works (e.g. \citealt{nalisnick2018deep,fetaya2019understanding,kirichenko2020normalizing}) that show that many modern generative models do not truly capture the underlying data distribution. As far as we know, GANs have not been analyzed through the LL, despite their ability to generate images perceptually similar to natural images, as there is no direct access to the model likelihood. 

There is a vast literature on evaluating whether GANs have learned the correct distribution (e.g. ~\citealt{heusel2017gans,sajjadi2018assessing,ravuri2019classification,webster2019detecting,naeem2020reliable,borji2022pros,ravuri2023understanding}). Previous works predominantly use samples from the GAN in order to evaluate them, which raises  some uncertainties because the space of all images is huge. In this work, we used the assumptions made when defining GANs in order to evaluate them, thereby side-stepping the issue of determining whether the learned distribution is correct through use of samples.

A key aspect of our investigation is the use of likelihood estimation and inference with GANs. There have been many works in the past that have suggested \emph{extensions} to GANs in order to allow them to work as classifiers or outlier detectors (e.g. \citealt{schlegl2017unsupervised,donahue2019large,kang2021rebooting,nitzan2022large}). Other works change the definition of GANs in order to train them as density estimators together with their adversarial loss (\citealt{lucas2019adaptive,abbasnejad2019generative}). All such methods use an additional component besides the GAN generator, such as an encoder, which is not part of the generative process of the GAN. 
In this work, we utilize the above mentioned tasks as a way to better analyze how well the GAN has captured the true data distribution. We argue that if GANs accurately capture the true data distribution, their generator \emph{alone} should be enough to achieve near-optimal performance in classification/outlier detection. Additionally, even with the added observation model, we should expect that the training images would be a part of the typical set of images supported by the GAN. 

%\todo{Is the following last paragraph ok?}
As mentioned in the introduction, GANs have been shown to not fully capture the training distribution in the past, especially due to ``mode collapse'' (\citealt{goodfellow2016nips,richardson2018gans}).~\citeauthor{arora2018gans} have quantified mode collapse by  a ``birthday paradox'' test, in which they found that the support of many GANs is smaller than the size of the training distribution. To a large part, mode collapse seems not to be a problem of modern GANs which capture the full diversity of the training distribution, as evidenced by excellent FID scores.
Instead, our analysis points to a deeper problem than previously investigated. Instead of fitting only small parts of the distribution, our results indicate that GANs misrepresent almost all parts of the distribution. 
%These findings are in contrast to the working assumptions made by many works, wherein the fact that GANs are able to imitate some statistics of natural images is taken as proof that GANs have learned to represent the full distribution.

\section{Discussion}

Modern GANs are able to generate incredibly realistic images, that seem to arise from the same distribution as the images used to train the GAN. Moreover, they have excellent performance under modern evaluation metrics.
These facts have led to a common belief that GANs capture the true data manifold and the true data distribution.

When analyzing GANs as manifold methods, we find that their manifolds do not fit the training samples. Quantitatively, this failure can be summarized by the classification and outlier detection performance of the GANs when defining the distance from the manifold as an estimate for whether the sample is from the GAN's distribution or not. Even when relaxing the requirement that samples should be part of the GAN manifold does not improve the situation, as is evidenced by the poor generative classification behavior. Worse, we showed how the training examples are not part of the typical set of the GAN.

Even though this work has presented an overall negative view of GANs, they are still extremely powerful data samplers. Indeed, this fact was taken advantage of in domains such as image manipulation, style transfer and data augmentation, to great effect. However, our work shows that a more cautious stance should be taken when attempting to use GANs as priors for the true data distribution.

\subsection*{Acknowledgements}
We thank the Gatsby Foundation for financial support.

\bibliography{main}

\begin{thebibliography}{50}
\providecommand{\natexlab}[1]{#1}
\providecommand{\url}[1]{\texttt{#1}}
\expandafter\ifx\csname urlstyle\endcsname\relax
  \providecommand{\doi}[1]{doi: #1}\else
  \providecommand{\doi}{doi: \begingroup \urlstyle{rm}\Url}\fi

\bibitem[Abbasnejad et~al.(2019)Abbasnejad, Shi, Hengel, and
  Liu]{abbasnejad2019generative}
M~Ehsan Abbasnejad, Qinfeng Shi, Anton van~den Hengel, and Lingqiao Liu.
\newblock A generative adversarial density estimator.
\newblock In \emph{Proceedings of the IEEE/CVF Conference on Computer Vision
  and Pattern Recognition}, pp.\  10782--10791, 2019.

\bibitem[Arora et~al.(2018)Arora, Risteski, and Zhang]{arora2018gans}
Sanjeev Arora, Andrej Risteski, and Yi~Zhang.
\newblock Do gans learn the distribution? some theory and empirics.
\newblock In \emph{International Conference on Learning Representations}, 2018.

\bibitem[Bendel et~al.(2022)Bendel, Ahmad, and Schniter]{bendel2022regularized}
Matthew Bendel, Rizwan Ahmad, and Philip Schniter.
\newblock A regularized conditional gan for posterior sampling in inverse
  problems.
\newblock \emph{arXiv preprint arXiv:2210.13389}, 2022.

\bibitem[Borji(2022)]{borji2022pros}
Ali Borji.
\newblock Pros and cons of gan evaluation measures: New developments.
\newblock \emph{Computer Vision and Image Understanding}, 215:\penalty0 103329,
  2022.

\bibitem[Brock et~al.(2018)Brock, Donahue, and Simonyan]{brock2018large}
Andrew Brock, Jeff Donahue, and Karen Simonyan.
\newblock Large scale gan training for high fidelity natural image synthesis.
\newblock \emph{arXiv preprint arXiv:1809.11096}, 2018.

\bibitem[Cover(1999)]{cover1999elements}
Thomas~M Cover.
\newblock \emph{Elements of information theory}.
\newblock John Wiley \& Sons, 1999.

\bibitem[Donahue \& Simonyan(2019)Donahue and Simonyan]{donahue2019large}
Jeff Donahue and Karen Simonyan.
\newblock Large scale adversarial representation learning.
\newblock \emph{Advances in neural information processing systems}, 32, 2019.

\bibitem[Dosovitskiy et~al.(2020)Dosovitskiy, Beyer, Kolesnikov, Weissenborn,
  Zhai, Unterthiner, Dehghani, Minderer, Heigold, Gelly,
  et~al.]{dosovitskiy2020image}
Alexey Dosovitskiy, Lucas Beyer, Alexander Kolesnikov, Dirk Weissenborn,
  Xiaohua Zhai, Thomas Unterthiner, Mostafa Dehghani, Matthias Minderer, Georg
  Heigold, Sylvain Gelly, et~al.
\newblock An image is worth 16x16 words: Transformers for image recognition at
  scale.
\newblock \emph{arXiv preprint arXiv:2010.11929}, 2020.

\bibitem[Duda et~al.(1973)Duda, Hart, and Stork]{duda1973pattern}
Richard~O Duda, Peter~E Hart, and David~G Stork.
\newblock \emph{Pattern classification and scene analysis}, volume~3.
\newblock Wiley New York, 1973.

\bibitem[Fetaya et~al.(2019)Fetaya, Jacobsen, Grathwohl, and
  Zemel]{fetaya2019understanding}
Ethan Fetaya, J{\"o}rn-Henrik Jacobsen, Will Grathwohl, and Richard Zemel.
\newblock Understanding the limitations of conditional generative models.
\newblock \emph{arXiv preprint arXiv:1906.01171}, 2019.

\bibitem[Goodfellow(2016)]{goodfellow2016nips}
Ian Goodfellow.
\newblock Nips 2016 tutorial: Generative adversarial networks.
\newblock \emph{arXiv preprint arXiv:1701.00160}, 2016.

\bibitem[Goodfellow et~al.(2014)Goodfellow, Pouget-Abadie, Mirza, Xu,
  Warde-Farley, Ozair, Courville, and Bengio]{goodfellow2014generative}
Ian Goodfellow, Jean Pouget-Abadie, Mehdi Mirza, Bing Xu, David Warde-Farley,
  Sherjil Ozair, Aaron Courville, and Yoshua Bengio.
\newblock Generative adversarial nets.
\newblock \emph{Advances in neural information processing systems}, 27, 2014.

\bibitem[Grosse et~al.(2015)Grosse, Ghahramani, and
  Adams]{grosse2015sandwiching}
Roger~B Grosse, Zoubin Ghahramani, and Ryan~P Adams.
\newblock Sandwiching the marginal likelihood using bidirectional monte carlo.
\newblock \emph{arXiv preprint arXiv:1511.02543}, 2015.

\bibitem[Heusel et~al.(2017)Heusel, Ramsauer, Unterthiner, Nessler, and
  Hochreiter]{heusel2017gans}
Martin Heusel, Hubert Ramsauer, Thomas Unterthiner, Bernhard Nessler, and Sepp
  Hochreiter.
\newblock Gans trained by a two time-scale update rule converge to a local nash
  equilibrium.
\newblock \emph{Advances in neural information processing systems}, 30, 2017.

\bibitem[Hinton et~al.(1995)Hinton, Dayan, Frey, and Neal]{hinton1995wake}
Geoffrey~E Hinton, Peter Dayan, Brendan~J Frey, and Radford~M Neal.
\newblock The" wake-sleep" algorithm for unsupervised neural networks.
\newblock \emph{Science}, 268\penalty0 (5214):\penalty0 1158--1161, 1995.

\bibitem[Kang et~al.(2021)Kang, Shim, Cho, and Park]{kang2021rebooting}
Minguk Kang, Woohyeon Shim, Minsu Cho, and Jaesik Park.
\newblock Rebooting acgan: Auxiliary classifier gans with stable training.
\newblock \emph{Advances in neural information processing systems},
  34:\penalty0 23505--23518, 2021.

\bibitem[Kang et~al.(2022)Kang, Shin, and Park]{kang2022studiogan}
Minguk Kang, Joonghyuk Shin, and Jaesik Park.
\newblock Studiogan: a taxonomy and benchmark of gans for image synthesis.
\newblock \emph{arXiv preprint arXiv:2206.09479}, 2022.

\bibitem[Kang et~al.(2023)Kang, Zhu, Zhang, Park, Shechtman, Paris, and
  Park]{kang2023scaling}
Minguk Kang, Jun-Yan Zhu, Richard Zhang, Jaesik Park, Eli Shechtman, Sylvain
  Paris, and Taesung Park.
\newblock Scaling up gans for text-to-image synthesis.
\newblock In \emph{Proceedings of the IEEE/CVF Conference on Computer Vision
  and Pattern Recognition}, pp.\  10124--10134, 2023.

\bibitem[Karras et~al.(2020{\natexlab{a}})Karras, Aittala, Hellsten, Laine,
  Lehtinen, and Aila]{karras2020training}
Tero Karras, Miika Aittala, Janne Hellsten, Samuli Laine, Jaakko Lehtinen, and
  Timo Aila.
\newblock Training generative adversarial networks with limited data.
\newblock \emph{Advances in neural information processing systems},
  33:\penalty0 12104--12114, 2020{\natexlab{a}}.

\bibitem[Karras et~al.(2020{\natexlab{b}})Karras, Laine, Aittala, Hellsten,
  Lehtinen, and Aila]{karras2020analyzing}
Tero Karras, Samuli Laine, Miika Aittala, Janne Hellsten, Jaakko Lehtinen, and
  Timo Aila.
\newblock Analyzing and improving the image quality of stylegan.
\newblock In \emph{Proceedings of the IEEE/CVF conference on computer vision
  and pattern recognition}, pp.\  8110--8119, 2020{\natexlab{b}}.

\bibitem[Karras et~al.(2021)Karras, Aittala, Laine, H{\"a}rk{\"o}nen, Hellsten,
  Lehtinen, and Aila]{karras2021alias}
Tero Karras, Miika Aittala, Samuli Laine, Erik H{\"a}rk{\"o}nen, Janne
  Hellsten, Jaakko Lehtinen, and Timo Aila.
\newblock Alias-free generative adversarial networks.
\newblock \emph{Advances in Neural Information Processing Systems},
  34:\penalty0 852--863, 2021.

\bibitem[Kingma et~al.(2021)Kingma, Salimans, Poole, and
  Ho]{kingma2021variational}
Diederik Kingma, Tim Salimans, Ben Poole, and Jonathan Ho.
\newblock Variational diffusion models.
\newblock \emph{Advances in neural information processing systems},
  34:\penalty0 21696--21707, 2021.

\bibitem[Kingma \& Welling(2013)Kingma and Welling]{kingma2013auto}
Diederik~P Kingma and Max Welling.
\newblock Auto-encoding variational bayes.
\newblock \emph{arXiv preprint arXiv:1312.6114}, 2013.

\bibitem[Kingma \& Dhariwal(2018)Kingma and Dhariwal]{kingma2018glow}
Durk~P Kingma and Prafulla Dhariwal.
\newblock Glow: Generative flow with invertible 1x1 convolutions.
\newblock \emph{Advances in neural information processing systems}, 31, 2018.

\bibitem[Kirichenko et~al.(2020)Kirichenko, Izmailov, and
  Wilson]{kirichenko2020normalizing}
Polina Kirichenko, Pavel Izmailov, and Andrew~G Wilson.
\newblock Why normalizing flows fail to detect out-of-distribution data.
\newblock \emph{Advances in neural information processing systems},
  33:\penalty0 20578--20589, 2020.

\bibitem[Kynk{\"a}{\"a}nniemi et~al.(2019)Kynk{\"a}{\"a}nniemi, Karras, Laine,
  Lehtinen, and Aila]{kynkaanniemi2019improved}
Tuomas Kynk{\"a}{\"a}nniemi, Tero Karras, Samuli Laine, Jaakko Lehtinen, and
  Timo Aila.
\newblock Improved precision and recall metric for assessing generative models.
\newblock \emph{Advances in Neural Information Processing Systems}, 32, 2019.

\bibitem[Lucas et~al.(2019)Lucas, Shmelkov, Alahari, Schmid, and
  Verbeek]{lucas2019adaptive}
Thomas Lucas, Konstantin Shmelkov, Karteek Alahari, Cordelia Schmid, and Jakob
  Verbeek.
\newblock Adaptive density estimation for generative models.
\newblock \emph{Advances in Neural Information Processing Systems}, 32, 2019.

\bibitem[Menon et~al.(2020)Menon, Damian, Hu, Ravi, and Rudin]{menon2020pulse}
Sachit Menon, Alexandru Damian, Shijia Hu, Nikhil Ravi, and Cynthia Rudin.
\newblock Pulse: Self-supervised photo upsampling via latent space exploration
  of generative models.
\newblock In \emph{Proceedings of the ieee/cvf conference on computer vision
  and pattern recognition}, pp.\  2437--2445, 2020.

\bibitem[Naeem et~al.(2020)Naeem, Oh, Uh, Choi, and Yoo]{naeem2020reliable}
Muhammad~Ferjad Naeem, Seong~Joon Oh, Youngjung Uh, Yunjey Choi, and Jaejun
  Yoo.
\newblock Reliable fidelity and diversity metrics for generative models.
\newblock In \emph{International Conference on Machine Learning}, pp.\
  7176--7185. PMLR, 2020.

\bibitem[Nalisnick et~al.(2018)Nalisnick, Matsukawa, Teh, Gorur, and
  Lakshminarayanan]{nalisnick2018deep}
Eric Nalisnick, Akihiro Matsukawa, Yee~Whye Teh, Dilan Gorur, and Balaji
  Lakshminarayanan.
\newblock Do deep generative models know what they don't know?
\newblock \emph{arXiv preprint arXiv:1810.09136}, 2018.

\bibitem[Nalisnick et~al.(2019)Nalisnick, Matsukawa, Teh, and
  Lakshminarayanan]{nalisnick2019detecting}
Eric Nalisnick, Akihiro Matsukawa, Yee~Whye Teh, and Balaji Lakshminarayanan.
\newblock Detecting out-of-distribution inputs to deep generative models using
  typicality.
\newblock \emph{arXiv preprint arXiv:1906.02994}, 2019.

\bibitem[Neal(2001)]{neal2001annealed}
Radford~M Neal.
\newblock Annealed importance sampling.
\newblock \emph{Statistics and computing}, 11:\penalty0 125--139, 2001.

\bibitem[Nitzan et~al.(2022)Nitzan, Gal, Brenner, and
  Cohen-Or]{nitzan2022large}
Yotam Nitzan, Rinon Gal, Ofir Brenner, and Daniel Cohen-Or.
\newblock Large: Latent-based regression through gan semantics.
\newblock In \emph{Proceedings of the IEEE/CVF Conference on Computer Vision
  and Pattern Recognition}, pp.\  19239--19249, 2022.

\bibitem[Ravuri \& Vinyals(2019)Ravuri and Vinyals]{ravuri2019classification}
Suman Ravuri and Oriol Vinyals.
\newblock Classification accuracy score for conditional generative models.
\newblock \emph{Advances in neural information processing systems}, 32, 2019.

\bibitem[Ravuri et~al.(2023)Ravuri, Rey, Mohamed, and
  Deisenroth]{ravuri2023understanding}
Suman Ravuri, M{\'e}lanie Rey, Shakir Mohamed, and Marc~Peter Deisenroth.
\newblock Understanding deep generative models with generalized empirical
  likelihoods.
\newblock In \emph{Proceedings of the IEEE/CVF Conference on Computer Vision
  and Pattern Recognition}, pp.\  24395--24405, 2023.

\bibitem[Richardson \& Weiss(2018)Richardson and Weiss]{richardson2018gans}
Eitan Richardson and Yair Weiss.
\newblock On gans and gmms.
\newblock \emph{Advances in neural information processing systems}, 31, 2018.

\bibitem[Sajjadi et~al.(2018)Sajjadi, Bachem, Lucic, Bousquet, and
  Gelly]{sajjadi2018assessing}
Mehdi~SM Sajjadi, Olivier Bachem, Mario Lucic, Olivier Bousquet, and Sylvain
  Gelly.
\newblock Assessing generative models via precision and recall.
\newblock \emph{Advances in neural information processing systems}, 31, 2018.

\bibitem[Sauer et~al.(2022)Sauer, Schwarz, and Geiger]{sauer2022stylegan}
Axel Sauer, Katja Schwarz, and Andreas Geiger.
\newblock Stylegan-xl: Scaling stylegan to large diverse datasets.
\newblock In \emph{ACM SIGGRAPH 2022 conference proceedings}, pp.\  1--10,
  2022.

\bibitem[Schawinski et~al.(2017)Schawinski, Zhang, Zhang, Fowler, and
  Santhanam]{schawinski2017generative}
Kevin Schawinski, Ce~Zhang, Hantian Zhang, Lucas Fowler, and Gokula~Krishnan
  Santhanam.
\newblock Generative adversarial networks recover features in astrophysical
  images of galaxies beyond the deconvolution limit.
\newblock \emph{Monthly Notices of the Royal Astronomical Society: Letters},
  467\penalty0 (1):\penalty0 L110--L114, 2017.

\bibitem[Schlegl et~al.(2017)Schlegl, Seeb{\"o}ck, Waldstein, Schmidt-Erfurth,
  and Langs]{schlegl2017unsupervised}
Thomas Schlegl, Philipp Seeb{\"o}ck, Sebastian~M Waldstein, Ursula
  Schmidt-Erfurth, and Georg Langs.
\newblock Unsupervised anomaly detection with generative adversarial networks
  to guide marker discovery.
\newblock In \emph{International conference on information processing in
  medical imaging}, pp.\  146--157. Springer, 2017.

\bibitem[Song et~al.(2020)Song, Sohl-Dickstein, Kingma, Kumar, Ermon, and
  Poole]{song2020score}
Yang Song, Jascha Sohl-Dickstein, Diederik~P Kingma, Abhishek Kumar, Stefano
  Ermon, and Ben Poole.
\newblock Score-based generative modeling through stochastic differential
  equations.
\newblock \emph{arXiv preprint arXiv:2011.13456}, 2020.

\bibitem[Theis et~al.(2015)Theis, Oord, and Bethge]{theis2015note}
Lucas Theis, A{\"a}ron van~den Oord, and Matthias Bethge.
\newblock A note on the evaluation of generative models.
\newblock \emph{arXiv preprint arXiv:1511.01844}, 2015.

\bibitem[Tipping \& Bishop(1999)Tipping and Bishop]{tipping1999principal}
Michael~E Tipping and Christopher~M Bishop.
\newblock Principal component analysis.
\newblock 1999.

\bibitem[Webster et~al.(2019)Webster, Rabin, Simon, and
  Jurie]{webster2019detecting}
Ryan Webster, Julien Rabin, Loic Simon, and Fr{\'e}d{\'e}ric Jurie.
\newblock Detecting overfitting of deep generative networks via latent
  recovery.
\newblock In \emph{Proceedings of the IEEE/CVF Conference on Computer Vision
  and Pattern Recognition}, pp.\  11273--11282, 2019.

\bibitem[Wu et~al.(2016)Wu, Burda, Salakhutdinov, and
  Grosse]{wu2016quantitative}
Yuhuai Wu, Yuri Burda, Ruslan Salakhutdinov, and Roger Grosse.
\newblock On the quantitative analysis of decoder-based generative models.
\newblock \emph{arXiv preprint arXiv:1611.04273}, 2016.

\bibitem[Xiao et~al.(2021)Xiao, Kreis, and Vahdat]{xiao2021tackling}
Zhisheng Xiao, Karsten Kreis, and Arash Vahdat.
\newblock Tackling the generative learning trilemma with denoising diffusion
  gans.
\newblock \emph{arXiv preprint arXiv:2112.07804}, 2021.

\bibitem[Zhang et~al.(2018{\natexlab{a}})Zhang, Zhang, Zhang, Liu, and
  Khurshid]{zhang2018deeproad}
Mengshi Zhang, Yuqun Zhang, Lingming Zhang, Cong Liu, and Sarfraz Khurshid.
\newblock Deeproad: Gan-based metamorphic testing and input validation
  framework for autonomous driving systems.
\newblock In \emph{Proceedings of the 33rd ACM/IEEE International Conference on
  Automated Software Engineering}, pp.\  132--142, 2018{\natexlab{a}}.

\bibitem[Zhang et~al.(2018{\natexlab{b}})Zhang, Isola, Efros, Shechtman, and
  Wang]{zhang2018unreasonable}
Richard Zhang, Phillip Isola, Alexei~A Efros, Eli Shechtman, and Oliver Wang.
\newblock The unreasonable effectiveness of deep features as a perceptual
  metric.
\newblock In \emph{Proceedings of the IEEE conference on computer vision and
  pattern recognition}, pp.\  586--595, 2018{\natexlab{b}}.

\bibitem[Zhao et~al.(2020)Zhao, Liu, Lin, Zhu, and Han]{zhao2020differentiable}
Shengyu Zhao, Zhijian Liu, Ji~Lin, Jun-Yan Zhu, and Song Han.
\newblock Differentiable augmentation for data-efficient gan training.
\newblock \emph{Advances in Neural Information Processing Systems},
  33:\penalty0 7559--7570, 2020.

\bibitem[Zimmermann et~al.(2021)Zimmermann, Schott, Song, Dunn, and
  Klindt]{zimmermann2021score}
Roland~S Zimmermann, Lukas Schott, Yang Song, Benjamin~A Dunn, and David~A
  Klindt.
\newblock Score-based generative classifiers.
\newblock \emph{arXiv preprint arXiv:2110.00473}, 2021.

\end{thebibliography}
\bibliographystyle{tmlr}

\appendix

\section{Brief Explanation of the AIS Algorithm}\label{a:ais}

Let $f(z)$ be a target un-normalized distribution. An AIS chain is defined by an initial distribution $Q_0(z)=q_0(z)/Z_0$ whose normalization coefficient is known, together with $T$ intermediate distributions $Q_1(z),\cdots,Q_T(Z)$ such that $Q_T(z)=q_T(z)/Z_T=f(z)/Z_T$. Each step of the chain further requires an MCMC transition operator $\mathcal{T}_t$ which keeps $Q_t(z)$ invariant, such as Hamiltonian Monte-Carlo (HMC). Beginning with a sample from the initial distribution $z_0\sim Q_0(z)$ and setting $w_0=1$, AIS iteratively carries out the following steps:
\begin{equation}
    w_t=w_{t-1}\cdot\frac{q_t(z_{t-1})}{q_{t-1}(z_{t-1})} \qquad\qquad z_t\sim \mathcal{T}_t\lrpar{z|z_{t-1}}
\end{equation}
The importance weights $w_T$ aggregated during the sampling procedure are an unbiased estimate of the ratio of normalizing coefficients, such that $\mathbb{E}\lrbra{w_T}=Z_T/Z_0$.

Given a sample $x$, the likelihood can be calculated through AIS by setting $f(z)=p(z)p_\theta(x|z)$ as the target distribution, such that $p_\theta(x)$ is the corresponding normalization constant $\mathcal{Z}_T$. In practice, the likelihood is estimated in log-space to avoid numerical difficulties such as underflows. Calculating the log of the importance weights as described above is straightforward, however \citeauthor{grosse2015sandwiching} have shown that doing so results in a stochastic lower bound of the log-likelihood. As the number of intermediate steps $T$ increases, this stochastic lower bound becomes tighter and converges to the true log-likelihood.

% \section{Splitting the Log-Likelihood}\label{a:LL-KL-rec}
% In the main body of text, an identity to the log-likelihood of an image $\log p_\theta(x)$ was presented:
% \begin{equation}\label{eq:a-ll-equiv}
%     \log p_\theta(x)=\mathbb{E}_{z|x}\lrbra{\log p_\sigma(x|\ G_\theta(z))} - D_\text{KL}\lrpar{p_\theta(z|x)\ ||\ p(z)}
% \end{equation}

% We present the derivation for this equivalence here. From the definition of the KL-divergence:
% \begin{align}
%     D_{\text{KL}}\left(p_\theta(z|x)||p\left(z\right)\right) & =\intop p_\theta(z|x)\log\frac{p_\theta(z|x)}{p\left(z\right)}dz\\
%  & =\intop p_\theta(z|x)\left[\log\frac{p_\theta(z|x)}{p\left(z\right)}+\log p_\theta(x)-\log p_\theta(x)\right]dz\\
%  & =\intop p_\theta(z|x)\log\frac{p_\theta(z,x)}{p\left(z\right)}dz-\log p_\theta(x)\\
%  & =\intop p_\theta(z|x)\log p_\theta(x|z)dz-\log p_\theta(x)\\
%  & =\mathbb{E}_{z|x}\left[\log p_\theta(x|z)\right]-\log p_\theta(x)
% \end{align}

% Finally, equation~\ref{eq:a-ll-equiv} is reached by exchanging the log-likelihood term and KL term between the RHS and LHS.

\section{Implementation Details}\label{a:details}

\subsection{AIS Details}\label{a:ais-details}
\begin{figure}
\begin{center}
    \includegraphics[width=.66\textwidth]{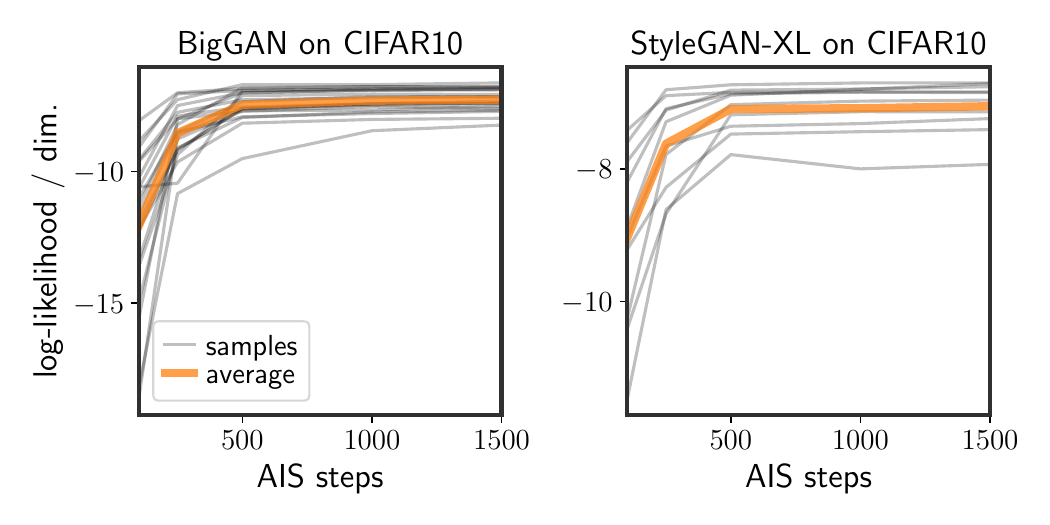}
\end{center}
  \caption{The estimated log-likelihood as a function of AIS steps, for different GANs on CIFAR10 on random test samples. The log-likelihood of most samples converges after 500 steps.}
  \label{fig:app-ais-steps}

\end{figure}
We follow the implementation of AIS from \citeauthor{wu2016quantitative} (publicly \href{https://github.com/tonywu95/eval_gen}{available in GitHub}), reimplemented in PyTorch. When possible, we used the same settings as ~\citeauthor{wu2016quantitative}: 
\begin{itemize}
    \item The transition operator we used was HMC with 10 leapfrog steps and a Metropolis-Hastings (MH) adjustment. During sampling, the learning rate is initialized to $5\cdot 10^{-2}$ and adjusted according to a moving average of the MH rejection rate
    \item During sampling, the intermediate distributions we used were:
    \begin{equation}
        Q_t(z)\propto p(z)\cdot p_\gamma\left(x|\ G_\theta(z)\right)^{\beta_t}
    \end{equation}
    $\beta_t$ was annealed according to a sigmoidal schedule
\end{itemize}

\paragraph{Choice of Number of Steps}
The bound on the log-likelihood approximated by AIS becomes tight and accurate only as the number of intermediate distribution and number of chains grows, respectively. However, AIS with many chains and intermediate distributions is incredibly computationally costly. Due to these considerations we use a relatively small number of intermediate distributions, while still ensuring accurate enough results.

Each chain used 500 intermediate steps. This number is in stark contrast to the 10,000 iterations used by \citeauthor{wu2016quantitative}. We chose this number by running multiple AIS chains with a differing number of intermediate steps, plotting the estimated log-likelihood as a function of AIS steps, as shown in Figure~\ref{fig:app-ais-steps}. The estimated log-likelihood typically stabilizes very close to the value reached after 500 iterations. The difference between the converged value and the one after 500 steps is much smaller than the resolution of log-likelihoods we are looking at, so this is a compromise between accuracy and computational cost.  

Further justification for this is due to the comparison between AIS and GAN inversion in terms of gradient steps. Because of the leapfrog steps, a single iteration of AIS is similar to 10 gradient steps in GAN inversion. In all of our experiments, $\sim$750 iterations were enough to converge during inversion, well below the 5000 gradient steps used during the AIS procedure.

\subsection{Reconstruction through GAN Inversion}\label{a:inversion-details}
There is a vast literature on the best way to reconstruct test images using GANs, also called \emph{GAN inversion}. In this work we used a simple, albeit rather costly, approach in order to find the best possible reconstruction. 

We used an optimization approach towards GAN inversion, using ADAM as the optimizer and a cosine schedule (similar to the scheme used by \citeauthor{sauer2022stylegan} in their implementation) for $\sim500-750$ iterations (which we found to usually be many more iterations than required). To find better reconstructions, we sampled $\sim500$ images from the GAN and initialized the optimizer from the latent code of the image closest to the input image in $\ell_2$ distance. Furthermore, this process was repeated $\sim4$ times for each image. Using this GAN inversion scheme, we were typically able to invert images generated by the GAN (and frequently images of a single color as well).

Finally, note that for all experiments with the StyleGAN variants, the inversion took place in $\mathcal{Z}$ space, as the generative model is defined in terms of this latent space and not the $\mathcal{W}/\mathcal{W+}$ spaces.

\subsection{Observation Model}\label{a:variance}

To calculate the GANs' density, we assumed an additive Gaussian noise observation model, with the same variance $\sigma^2$ for all samples. This observation model was used by~\citeauthor{wu2016quantitative} among others. The values of $\sigma^2$ and their respective PSNR can be found in Table~\ref{tab:gamma-vals}. As mentioned in the main body of text, the value of $\sigma^2$ used was the variance of the GAN's reconstruction error. This choice of $\sigma^2$ maximizes the likelihood the model gives to training data.

\begin{table}[h]
    \centering

    \begin{tabular}{|c|c|c|c|} \hline  
         GAN&  Dataset&  Value of $\sigma^2$& PSNR\\ \hline  \hline
         BigGAN&  CIFAR10&  0.008& 21.2\\ \hline  
         StyleGAN-XL&  CIFAR10&  0.018& 17.4\\ \hline  \hline
         BigGAN&  ImageNet10&  0.012& 19.2\\ \hline
         StyleGAN-XL& ImageNet10& 0.021&16.8\\\hline \hline
         StyleGAN2-ADA& AFHQ& 0.019&17.2\\\hline
    \end{tabular}    
    \caption{\label{tab:gamma-vals}Table of variances used for $\sigma^2$ throughout the paper and their respective PSNR.}
\end{table}

\section{More Results}\label{a:results}

\begin{figure}
\begin{center}
    \includegraphics[width=\linewidth]{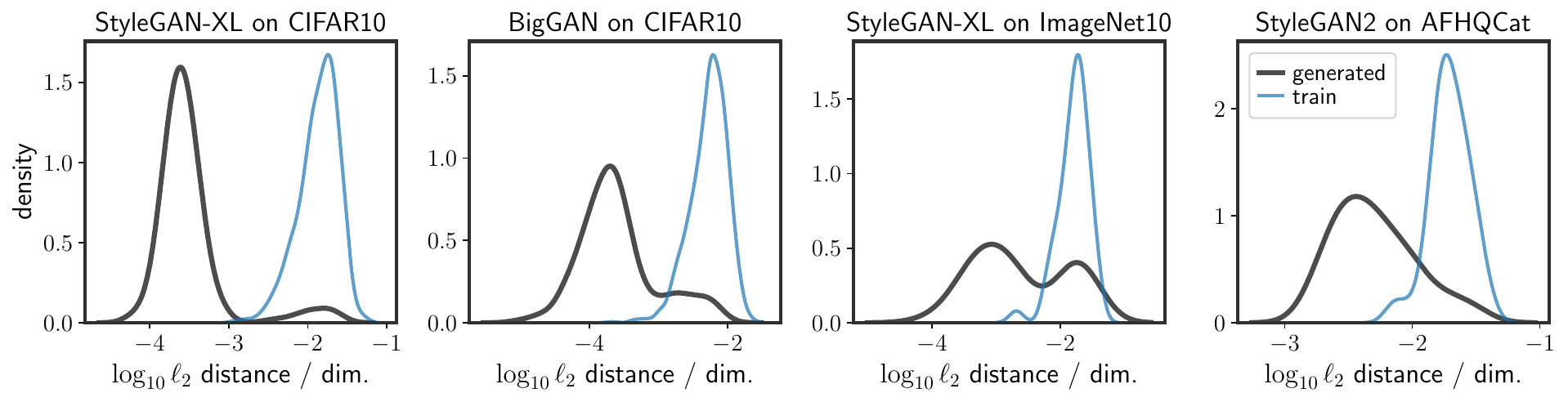}
\end{center}
  \caption{Log reconstruction errors of training vs generated examples for more GANs. In all cases, the distribution of the training examples is different than of the generated examples. The plot for StyleGAN-XL on ImageNet is on a bigger subset of ImageNet classes than in figure~\ref{fig:train-recon}, showing that this result is broader than the 2 classes discussed in the main body of text.}
  \label{fig:app-inv-vs-gen}

\end{figure}
\begin{figure}
\begin{center}
    \includegraphics[width=\linewidth]{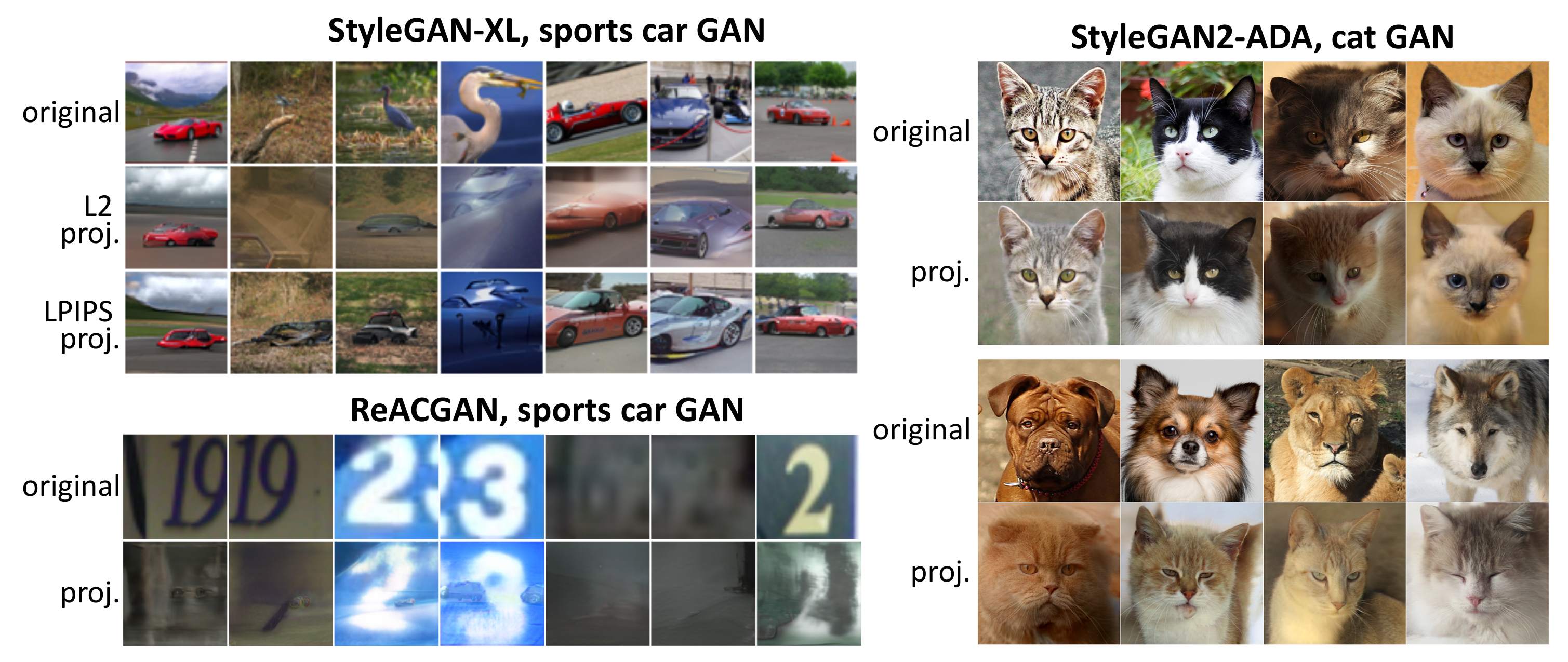}
\end{center}
  \caption{More GAN inversions.}
  \label{fig:app-invs}

\end{figure}
\begin{figure}
\begin{center}
    \includegraphics[width=\linewidth]{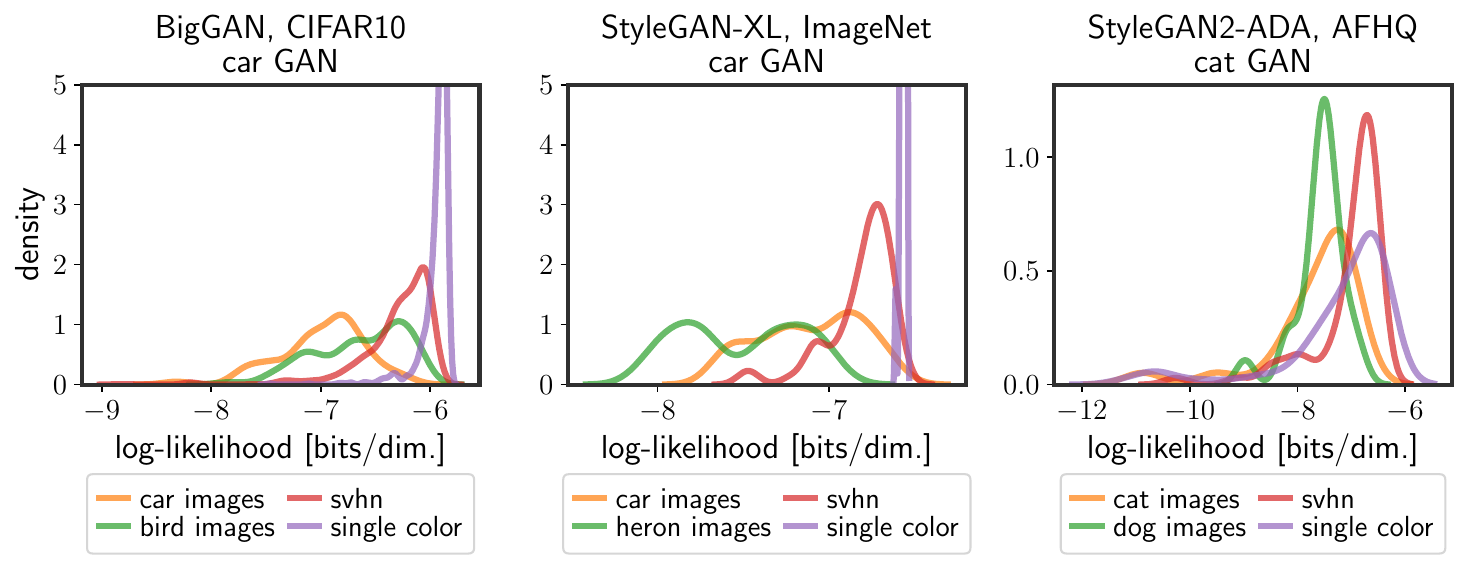}
\end{center}
  \caption{Histograms of LLs for more GANs than shown in the main text.}
  \label{fig:app-LL-hists}

\end{figure}

Figure~\ref{fig:app-inv-vs-gen} shows more histograms of the reconstruction errors for training examples vs. images generated by GANs. In all cases, the distribution of the reconstruction errors is different for train and generated examples, indicating that for all GANs examined the training samples are not on the GAN manifold.

Figure~\ref{fig:app-invs} shows more examples of reconstructions. 

Figure~\ref{fig:app-LL-hists} shows LL histograms for different image groups under different GANs.

\end{document}